\documentclass[letterpaper, 10 pt, conference]{ieeeconf}  

\IEEEoverridecommandlockouts                              
\usepackage[utf8]{inputenc}
\usepackage[T1]{fontenc}
\usepackage{graphicx}
\usepackage[namelimits]{amsmath} 
\usepackage{amssymb}             
\usepackage{amsfonts}            
\usepackage{mathrsfs}            
\usepackage{subfigure} 
\usepackage{booktabs}
\usepackage{cite}
\usepackage{url}
\usepackage{dsfont}
\usepackage{algorithm}
\usepackage{algpseudocode}
\usepackage{algorithmicx}
\usepackage{amsmath}
\usepackage[switch]{lineno}
\usepackage{bbding}

\usepackage{hyperref}

\usepackage{multirow}
\usepackage{textcomp,booktabs}
\usepackage[usenames,dvipsnames]{color}
\usepackage{colortbl}
\definecolor{mgray}{gray}{.9}
\hypersetup{
    colorlinks=true,
    linkcolor=blue,
    filecolor=magenta,      
    urlcolor=cyan
}

\usepackage[colorinlistoftodos,bordercolor=blue,backgroundcolor=blue!20,linecolor=blue,textsize=scriptsize]{todonotes}
\title{\LARGE \bf
Occupancy Prediction-Guided Neural Planner for Autonomous Driving}

\author{Haochen Liu, Zhiyu Huang, and Chen Lv$^{*}$,~\IEEEmembership{Senior Member, IEEE}
\thanks{Code is available at: {\href{https://github.com/georgeliu233/OPGP}{https://github.com/georgeliu233/OPGP}}}
\thanks{H. Liu, Z. Huang, and C. Lv are with the School of Mechanical and Aerospace Engineering, Nanyang Technological University, 639798, Singapore. (E-mails: {\tt \{haochen002, zhiyu001\}@e.ntu.edu.sg, lyuchen@ntu.edu.sg})}
\thanks{This work was supported in part by the A*STAR under MTC IRG Grant (No. M22K2c0079) and the SUG-NAP Grant (No. M4082268.050) of Nanyang Technological University, Singapore.}
\thanks{$^{*}$Corresponding author: C. Lv}
}

\begin{document}
\maketitle
\thispagestyle{empty}
\pagestyle{empty}

\begin{abstract}
Forecasting the scalable future states of surrounding traffic participants in complex traffic scenarios is a critical capability for autonomous vehicles, as it enables safe and feasible decision-making. Recent successes in learning-based prediction and planning have introduced two primary challenges: generating accurate joint predictions for the environment and integrating prediction guidance for planning purposes. To address these challenges, we propose a two-stage integrated neural planning framework, termed OPGP, that incorporates joint prediction guidance from occupancy forecasting. The preliminary planning phase simultaneously outputs the predicted occupancy for various types of traffic actors based on imitation learning objectives, taking into account shared interactions, scene context, and actor dynamics within a unified Transformer structure. Subsequently, the transformed occupancy prediction guides optimization to further inform safe and smooth planning under Frenet coordinates. We train our planner using a large-scale, real-world driving dataset and validate it in open-loop configurations. Our proposed planner outperforms strong learning-based methods, exhibiting improved performance due to occupancy prediction guidance.
\end{abstract}


\section{Introduction}
\label{sec1}
Accurately predicting robust, socially compatible joint futures for multiple traffic participants (agents) to inform the decision-making module is a crucial capability for autonomous driving systems (ADS) \cite{mozaffari2020deep, chen2022milestones, badue2021self}. However, integrating prediction and planning presents significant challenges due to several factors. First, ADS must process a vast array of complex traffic scenes, each featuring combinations of socially interactive heterogeneous traffic participants \cite{wang2022social}. Second, the motion predictor must manage joint patterns of future states for numerous traffic actors in the vicinity \cite{mo2020interaction}. Moreover, path-level planning decisions necessitate feasible guidance from joint predictions to execute safe and smooth planning performances for ADS.

To achieve optimal prediction guidance, the majority of prediction research focuses on multi-agent trajectory prediction (MATP), which directly maps the joint sequences of future locations for selected surrounding participants \cite{gu2021densetnt, salzmann2020trajectron++}. However, this approach is hindered by the search cost, which grows linearly with the number of participants and exponentially for marginal motion predictions \cite{huang2023gameformer}. Recently, researchers have turned to forecasting occupancy grids \cite{mahjourian2022occupancy}.

\begin{figure}[ht]
    \centering
    \includegraphics[width=\linewidth]{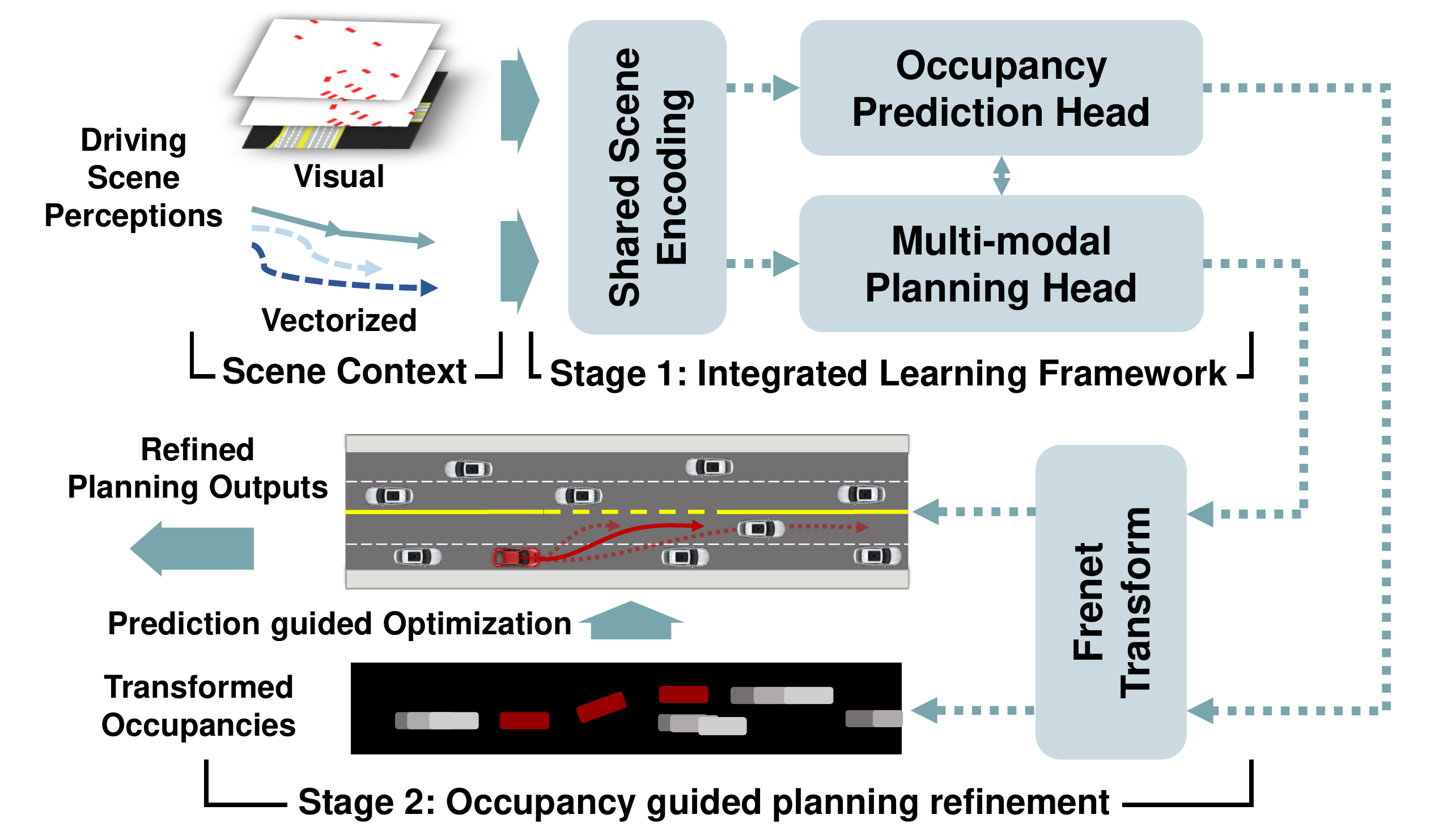}
    \caption{Proposed two-stage occupancy prediction-guided neural planner (OPGP). Predicted occupancy and multi-modal planning outputs from an integrated learning-based framework. We then refine the planning from transformed prediction and planning via prediction-guided optimizations.}
    \label{fig1}
\end{figure}

As a visual representation, occupancy prediction provides a more efficient and scalable form for multiple participants. In addition, it can predict the joint future for various participants at a single glance \cite{kim2022stopnet}, offering enhanced safety for occlusion predictability compared to MATP when integrating planning decisions. However, capturing the social awareness of all interactive actors in the context of joint occupancy forecasting remains an open question. 

Conversely, learning-based motion planning has gained increasing attention \cite{zhou2022review}. Its ability to handle diverse driving scenarios has achieved impressive results through the use of a unified neural network. Nevertheless, the robustness and safety of learning-based planning can be undermined by instability. To address this issue, integrating prediction guidance has emerged as a standard approach for this learning structure \cite{aradi2020survey,liu2022augmenting, ngiam2021scene}. By learning active responses and motion constraints based on neighboring predictions, a safer, more robust, and socially compliant planning system is anticipated \cite{espinoza2022deep}. One common method involves integrating predictions by using the future trajectories of surrounding actors for planning \cite{huang2022differentiable,cui2021lookout}. Despite the difficulties of using MATP to model joint trajectories, limited research has explored occupancy as a source of prediction guidance. Predicting joint future occupancy presents considerable challenges in integrating visual-based prediction features within the neural network for planning. Furthermore, transforming informative predicted occupancy into feasible planning outcomes is an issue that warrants further investigation.

To tackle these challenges, as shown in Fig. \ref{fig1}, we propose a two-stage learning-based framework (named OPGP) that integrates joint predictions for future occupancy and motion planning with prediction guidance. In the first stage, an integrated network of occupancy prediction-guided planning is established upon Transformer backbones. Building upon our previous work \cite{liu2022strajnet}, occupancy predictions for all types of traffic participants are output simultaneously, taking into account interaction awareness for both visual features and vectorized context. Meanwhile, encoded scene features and occupancy are shared and conditionally queried in the planner head, which conducts multi-modal motion planning for the ADS's trajectories. The second stage focuses on modeling explicit guidance from occupancy prediction for motion planning refinement in an optimization-feasible manner. More specifically, we construct an optimization pipeline in Frenet space \cite{werling2010optimal} for planning refinement using transformed occupancy predictions. The contributions of our proposed framework are summarized as follows:

\begin{enumerate}
\item We propose a two-stage occupancy prediction guided neural planner pipeline. The first stage involves an integrated occupancy prediction-guided motion planning framework. Predicted occupancy is combined with multi-modal planning trajectories in Transformer-based structures, taking into account the social interactions among traffic participants and scene context.

\item In the second stage, we design a transformed occupancy prediction guided optimization for planning refinement to further enhance planning performance.

\item We validate the two-stage framework on a large-scale real-world driving dataset, and the proposed pipeline achieves compelling performance.
\end{enumerate}

\section{Related Work}
\subsection{Joint motion prediction}
A growing number of learning-based techniques have been developed for joint motion predictions, proving highly effective \cite{mozaffari2020deep}. This can be attributed to the capability of deep neural networks, particularly Transformers and GNNs \cite{velivckovic2017graph}, to handle intricate traffic scenarios involving multiple interacting participants and diverse scene contexts. Agent-centric methods predict multi-agent-wise future trajectories (MATP) anchored on each detected traffic participant. DenseTNT \cite{gu2021densetnt}, M2I \cite{sun2022m2i}, and HEAT \cite{mo2022multi} score joint predictions in combinations of marginal ones for each actor with GNN backbones. While they demonstrate accuracy for each actor, they also raise computational costs with occasional inconsistency. Joint methods \cite{gilles2022gohome,gilles2021thomas} directly estimate the future distribution of all actors from heatmap scoring or sampling. However, these methods require a maximum number of predictions and have linear cost growth with multiple agents, especially in crowded urban areas. The use of occupancy grids to forecast future motions has been a technique employed in autonomous driving for several years. One notable example is ChauffeurNet \cite{bansal2018chauffeurnet}, which utilizes occupancy maps to predict future movements and facilitate behavior planning. StopNet \cite{kim2022stopnet} further conducts integrated trajectory and occupancy predictions for scalable and real-time predictions. Building on our previous method \cite{liu2022strajnet}, we incorporate planning information into the learning framework during prediction training. We have refined our occupancy prediction pipeline to include all types of traffic participants and consider their interactions within the scene context for enhancement to guide planning.

\subsection{Motion planning with prediction guidance}
Motion planning is a well-established field that has received extensive study over time. It has developed through various approaches such as path optimization \cite{hang2020integrated}, sampling \cite{huang2021driving}, and more recently, learning-based techniques \cite{gonzalez2015review,zhou2022review}. However, cruising safely in interactive and sophisticated traffic requires further guidance from the predicted states of participants' behaviors. PiP \cite{song2020pip} makes conditional predictions iteratively on sampling-based planning rescheduling, which only considers marginal futures and is limited by generated planning paths. DIPP \cite{huang2022differentiable} ties differentiable planning objectives with joint trajectory predictions, enabling responsive planning. While previous methods focus on prediction guidance in an agent-centric manner, we seek scene-centric guidance in terms of scalability and invariance for growing actors. PredictionNet \cite{kamenev2022predictionnet} demonstrates occupancy prediction guidance with post-processed trajectories to inform planning and control. However, it focuses on occupancy prediction for traffic simulation and does not specifically design integration for planning. End-to-end methods such as MP3 \cite{casas2021mp3} and Interfuser \cite{shao2023safety} directly map prediction and planning from raw visual inputs. Nevertheless, they focus more on unifying the perception part into prediction, and the planning outputs are simply retrieved from extensive trajectory storage. In our work, we leverage occupancy prediction to guide both stages of our proposed neural planner. Specifically, the prediction features are queried interactively by the learning-based planning decoder, and the resulting predictions are utilized to refine the planning in a feasible manner.

\section{Methodology}
\subsection{Problem Formulation}

\begin{figure*}[ht]
    \centering
    \includegraphics[width=0.94\linewidth]{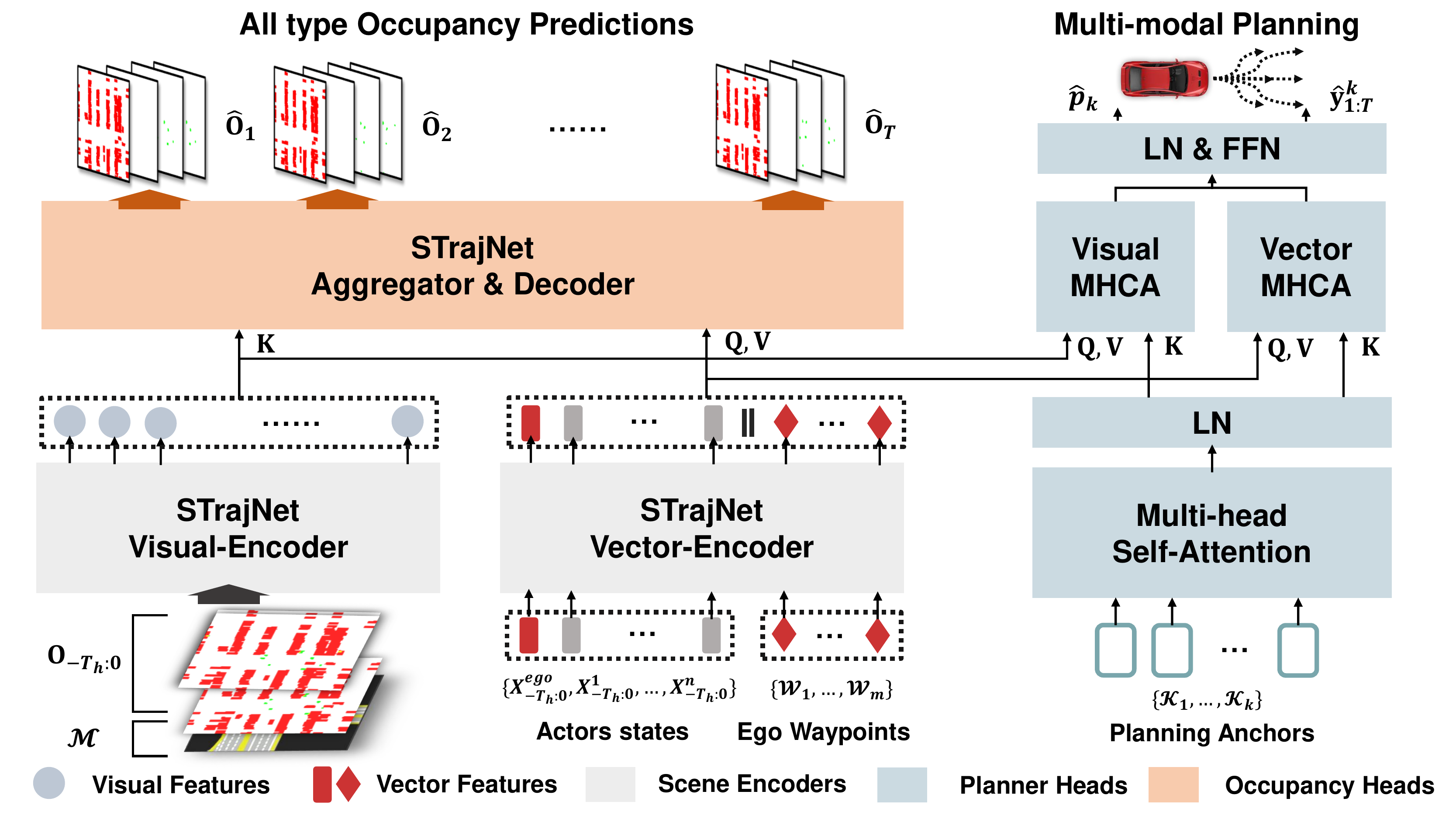}
    \caption{An overview of the integrated occupancy prediction and planning learning framework in OPGP. Building on our previous work, visual and vector scene inputs $\mathcal{S}$ are separately encoded in scene encoders; the encoded scene features are distributed by occupancy prediction heads for all type occupancy predictions $\mathbf{\hat{O}}_{1:T}$ and multi-modal planning heads for initial planning outputs $\mathbf{\hat{Y}}^e_{1:T}$ for the subsequent  stage.}
    \vspace{-0.1cm}
    \label{fig2}
\end{figure*}

As shown in Fig. \ref{fig1}, the occupancy prediction guided neural planner consists of two stages. In the first stage, motion-planning decisions with occupancy prediction guidance can be formulated as a multi-task learning paradigm. Conditioned on scene context $\mathcal{S}$ as inputs, the integrated learning framework $f$ simultaneously outputs occupancy predictions $\mathbf{\hat{O}}{1:T}$ and multi-modal imitated planning trajectories $\mathbf{\hat{Y}}^e{1:T}$ in future horizons $T$. Subsequently, the second stage utilizes a Frenet transformation \cite{werling2010optimal} for both prediction and planning to achieve attainable prediction-guided optimization that generates refined planning outputs $\tau^*$, minimizing the defined costs $C$. Mathematically, the two-stage task is formulated as:
\begin{equation}
\label{e2}
\begin{aligned}
\mathbf{\hat{Y}}^e_{1:T}, \mathbf{\hat{O}}_{1:T} &= f(\mathcal{S}|\theta), \\
\tau^*=\mathop{\arg\min}_{\tau} \ C(&\mathbf{\hat{Y}}^\mathbf{T}_{1:T}, \mathbf{\hat{O}}^\mathbf{T}_{1:T}, \mathcal{S}),
\end{aligned}
\vspace{-0.1cm}
\end{equation}
where $\theta$ denotes the model parameters. More specifically, the input scene context $\mathcal{S}$ comprises multiple modalities under historical horizon $T_h$ with detailed formulations as follows: \textbf{1) Visual features:} We construct a combination from the historical occupancy grids $\mathbf{O}_{T_h:0}$ and rasterized BEV roadmap $\mathcal{M}$ \cite{huang2022recoat} representing the spatial-temporal status of traffic participants under a specific scenario. Each occupancy $\mathbf{O}_t$ separately contains 3 types of traffic participants (vehicles, pedestrians, and cyclists), and we further add an occlusion dimension for predictions $\mathbf{\hat{O}}_{1:T}$ for currently unseen actors. \textbf{2) Vector features:} The vectorized context initially concerns the dynamic context from a maximum of $n_x+1$ actors centered on the autonomous vehicle $\mathbf{X}_{-T_h:0} = {X^{ego}_{-T_h:0}, X^1_{-T_h:0}, \cdots, X^n_{-T_h:0}}$, each representing a historical motion sequence: $(x, y, v_x, v_y, \theta)$. One-hot encoding is associated with respective types.  To provide map context for the autonomous (ego) vehicle, we collect map data using a graph search of Lanelets \cite{bender2014lanelets}, which includes $m$ waypoints $\mathcal{W}_m$ and reference routes $\mathcal{I}$. Each map segment contains coordinate information as well as road features such as road type, speed limit, and the current state of nearby traffic lights.

\subsection{Integrated Prediction and Planning Framework}
Fig. \ref{fig2} illustrates the overall structure of our learning-based framework, which integrates occupancy predictions and planning and consists of three modules. First, we adopt the separate encoding for multi-modal scene inputs $\mathcal{S}$ from our previous work. Encoded visual and vector features are subsequently delivered to occupancy prediction and planner heads. For occupancy heads, we maintain the previous aggregation and decoding pipelines, which then conduct a late fusion of encoded vector and visual features. The prediction heads ultimately output all types of occupancy predictions $\mathbf{\hat{O}}{1:T}$ for each second of the future horizon. For multi-modal planning heads, we follow the Transformer decoder paradigm and employ a single-layer decoding head for multi-modal path planning $\mathbf{\hat{Y}}^e{1:T}$. More specifically, we initially introduce a set of $k$ planning anchors ${\mathcal{K}_1,\cdot,\mathcal{K}_k}$ to guide $k$ modes for planning heads. Each anchor $\mathcal{K}$ can be either static end-point clusters or dynamic ones for learnable embedding that heuristically guide the planning decoding. These anchors then serve as context queries $\textbf{Q}$ for encoded visual and vector features after self-attentions.

For the multi-head cross-attention module (MHCA), we aim to adaptively obtain interaction-aware features from visual context, vectorized dynamics, and scene topology for the ego vehicle. We implement separate queries using different MHCA blocks for visual and vector features, serving as $\textbf{K}, \textbf{V}$. This is due to the length of padded visual features that will sparse the attention scoring when considering concatenated attention heads. After subsequent feed-forward networks (FFNs), each mode will decode to an output of a planning path $\hat{\mathbf{y}}^k_{1:T}$ and corresponding probabilities $\hat{p}_k$ for imitation likelihoods.

We formulate the first-stage learning framework as a multi-task learning paradigm, and the objectives are offline updated jointly using large-scale datasets. For occupancy predictions, as each grid cell forecasts the occupying probabilities under a certain future horizon, we employ focal loss \cite{lin2017focal} to balance the occupied samples for each type of actor:
\begin{equation}
    \vspace{-0.1cm}
    \mathcal{L}_{pred} = \frac{1}{HWTU}\sum_{x,y,t,u}^{H,W,T,U}\operatorname{FL}(\mathbf{\hat{O}}^u_t(x,y),\mathbf{O}^u_t(x,y)),
    \vspace{-0.1cm}
\end{equation}
where $H,W$ represent the occupancy scales and $U$ denotes the actor types. For multi-modal planning, the imitation learning objectives are updated as follows:
\begin{equation}
    \vspace{-0.2cm}
    \mathcal{L}_{plan} = \mathop{\arg\min}_{k}\frac{1}{T}\sum_t^T\operatorname{SL_1}(\hat{y}^k_t - y^k_t) - p_k\log{\hat{p}_k},
\end{equation}
where $\operatorname{SL_1}$ denotes smooth L1 loss, and the planning modes with minimum distance to ground truth will be updated for the initial planning path $\hat{\mathbf{Y}}^e_{1:T}$.

\subsection{Prediction-guided planning refinement}
Given the occupancy prediction for scene traffic participants $\mathbf{\hat{O}}_{1:T}$ as well as the multi-modal initial planning results $\hat{\mathbf{Y}}^e_{1:T}$, the second stage of OPGP constructs a prediction-guided pipeline for planning refinement. First, this section designs a transformation paradigm using Frenet \cite{werling2010optimal} coordinates for both prediction and planning. Then, we introduce a planning optimization guided by the transformed occupancy predictions.

\textbf{Transformation:} We leverage Frenet coordinates for transformation targets, as they relax the difficulties in planning optimizations. Suppose a generated reference route $\mathcal{I}$, each reference point $r \in \mathcal{I}$ is dynamically assigned with a curvilinear frame by tangential and normal vectors: $[\Vec{t}_r, \Vec{n}_r]$. Then, the current Cartesian coordinate $\Vec{y}=(x,y)$ can be transformed into Frenet $\Vec{r}=(s,d)$ through the transformation $\mathcal{F}:\Vec{y}\rightarrow\Vec{r}$ following the relations:
\begin{equation}
\Vec{y}(s(t),d(t))=\Vec{r}(s(t)) + d(t)\Vec{n}_r(s(t)).
\end{equation}

For initial planning results $\hat{\mathbf{y}}^k_{1:T}$, we first select the top-scoring trajectories $\hat{\mathbf{Y}}^e_{1:T}=\mathop{\arg\max}_{p_k}\hat{y}^k{1:T}$, and then transform them by $\hat{\mathbf{Y}}^\mathbf{T}_{1:T}=\mathcal{F}(\hat{\mathbf{Y}}^e_{1:T})$. For predicted occupancy $\mathbf{\hat{O}}_{1:T}$, our objective is to filter out the occupancy prediction centered on $\mathcal{I}$ and stretch onto occupancy under the Frenet frame. This transformation focuses the searching space of scalable occupancy on planning without occlusions. Moreover, the Frenet-transformed occupancy relieves the convexity for planning optimization. More specifically, given predefined mesh-grid indices on the Frenet frame: $\mathbf{I}_{sd}, s\in[0,S] ; d \in [-\frac{D}{2},\frac{D}{2}]$, this procedure transforms reversely from Frenet firstly onto the Cartesian frame by $\mathcal{F}^{-1}$. Then, following the relations from Cartesian to occupancy grids pixel $(w, h)$ denoted as $\mathcal{P}$:
\vspace{-0.2cm}
\begin{equation}
    (w, h) = \operatorname{int}(\frac{1}{n}(x - x^e_0, y - y^e_0)),
\end{equation}
where $\operatorname{int}$ is the rounding function and $n$ represents pixels per meter. The transformed occupancy prediction $\mathbf{\hat{O}}^{\mathbf{T}}{1:T}$ is then gathered by:
\begin{equation}
\begin{aligned}
    \vspace{-0.2cm} 
    \mathbf{\hat{O}}^{\mathbf{T}}_{1:T}(s, d) &= \mathbf{\hat{O}}_{1:T}(\mathbf{I}_w, \mathbf{I}_h), \\
    [\mathbf{I}_w;\mathbf{I}_h] &= \mathcal{P}(\mathcal{F}^{-1}(\mathbf{I}_{sd})).
\end{aligned}
\end{equation}

In OPGP pipelines, the output transformed planning $\hat{\mathbf{Y}}^\mathbf{T}_{1:T}$ will then be optimized with the guidance of predictions from $\mathbf{\hat{O}}^{\mathbf{T}}_{1:T}$ for planning refinement.

\textbf{Optimization:} To ensure enhanced safety and motion performance for planning results, here we formulate an open-loop optimization problem under finite horizons. The optimization searches for an optimal sequence $\tau^*=\{\tau_1,\cdots,\tau_T\}$ that minimizes the cost function sets $C$ guided by transformed occupancy predictions $\mathbf{\hat{O}}^{\mathbf{T}}_{1:T}$. More specifically, $C$ is the weighted $\omega_i$ squared sum of a set of cost functions $c_i$:
\begin{equation}
\begin{aligned}
    \vspace{-0.2cm}
    \tau^*=&\mathop{\arg\min}_{\tau} C(\tau, \mathbf{\hat{O}}^\mathbf{T}_{1:T}, \mathcal{S}), \\
    C = \sum_i\Vert &\omega_i c_i(\tau^i,\mathbf{\hat{O}}^\mathbf{T}_{1:T}, \mathcal{S}) \Vert^2, \tau^i\subseteq\tau.
\end{aligned}
\label{equ7}
\end{equation}
The cost function set $C$ embraces various cost terms $c_i$ with carefully-devised weights $\omega_i$ that consider a variety of planning performances, including driving progress, driving comfort, adherence to rules, and most importantly, driving safety. The cost set is explicitly cataloged as follows:

\textbf{1) Driving progress:} To encourage efficient driving, we assist a consistent longitudinal speed cost within speed limits $v^s_{\operatorname{limit}}$, allowing more on-road progress:
\begin{equation}
    \textbf{c}^{\operatorname{progress}}_t = \Dot{s}_t - v^s_{\operatorname{limit}}.
\end{equation}

\textbf{2) Driving comfort:} To facilitate planning comfort and driving smoothness, we introduce the minimization for accelerations and jerks:
\begin{equation}
    \textbf{c}^{\operatorname{comfort}}_t =  \Ddot{s}_t + \Ddot{d}_t + \dddot{s_t}.
\end{equation} 

\textbf{3) Adherence to rules:} We expect the designed ADS to conform to traffic rules and stay within reference lanes. Thus, we propose an off-route cost and a traffic light cost based on max-margin objectives on $(s,d)$ respectively:
\begin{equation}
    \textbf{c}^{\operatorname{route}}_t = d_t.
\end{equation} 
\begin{equation}
    \textbf{c}^{\operatorname{tl}}_t = \left\{
\begin{aligned}
& \mathbf{1}(s_{\operatorname{red}})(s_t -  s_{\operatorname{red}}), \ s_t > s_{\operatorname{red}}, \\
& 0,                                                                 \ \operatorname{otherwise}.
\end{aligned}
\right.
\end{equation}
where $s_{\operatorname{red}}$ is the stop point for red-light reference lane.

\textbf{4) Prediction guided safety:} To enhance driving safety for planning refinement, we leverage the guidance of transformed occupancy predictions $\mathbf{\hat{O}}^{\mathbf{T}}_{1:T}$ as dynamic obstacles for safe driving. The occupancy prediction is considered as a potential collision cost weighted by actor types:
\begin{equation}
    \vspace{-0.2cm}
    \textbf{c}^{\operatorname{ogm}}_t(s,d) = \sum_u^{U}\lambda_u\mathbf{\hat{O}}^{\mathbf{T}u}_{t}(s,d),
\end{equation} 
Due to the issues of preserved occupancy between frames, the predicted actors in occupancy are possibly enlarged with long tails. This may lead to over-conservative driving for ADS. Therefore, we design a confidence safety distance $s^{\operatorname{safe}}_t$ for the confident collision cost over threshold $\epsilon$:
\begin{equation}
    \vspace{-0.2cm}
    s^{\operatorname{safe}}_t = \min_{s}(\sum_d^{D}\textbf{c}^{\operatorname{ogm}}_t(s,d) > \epsilon),
\end{equation} 

For optimization simplicity, we only conduct longitudinal guidance for planning refinement. Considering the forward sign distance:
\begin{equation}
    \operatorname{sgnd}(s_t) = (s_t - \mathbf{I}_s)\mathbf{1}(s_t > \mathbf{I}_s),
\end{equation} 

The prediction-guided safety cost is modeled as a weighted predicted collision cost sum for sign distance that surpasses $s^{\operatorname{safe}}_t$:
\begin{equation}
    \textbf{c}^{\operatorname{safe}}_t = 
\sum_{s}^{S}\sum_{d}^{D}\operatorname{sgnd}(s_t)\textbf{c}^{\operatorname{ogm}}_t(s,d)\mathbf{1}(s > s^{\operatorname{safe}}_t).
\end{equation}

\begin{table*}[htp]
\centering
\caption{Open-loop Testing Results on Waymo Motion Dataset}
\setlength{\tabcolsep}{4.2mm}{
\begin{tabular}{l|lll|lll|lll}
\toprule
 & \multicolumn{3}{c}{Safety (\%)}    & \multicolumn{3}{c}{Motion ($\operatorname{ms}^{-2},\operatorname{ms}^{-3}$)} & \multicolumn{3}{c}{Planning errors (m)} \\\midrule 
Models                        & Collisions & Off route & Red light & Jerk       & Acc.       & Lat.Acc.      & 1s          & 3s          & 5s          \\\midrule
Vanilla-IL              & 7.575      & 2.76      & 1.671     & 4.323      & 0.676      & 0.129         & \textbf{0.195}       & \textbf{1.076}       & \textbf{2.759}       \\
DIM-OPGP\cite{rhinehart2018deep}   & 9.275      & 14.27     & 3.246     & 7.226      & 0.977      & 0.252         & 0.407       & 1.865       & 4.473       \\
Vanilla-Test & - & - & - &4.02 & 0.692 & 0.167 & - & - & - \\\midrule \rowcolor{mgray}
\textbf{OPGP}                       & \textbf{3.018}      & \textbf{1.809}     & \textbf{0.501}         & 3.869      & 0.551      & 0.156         & 0.226       & 1.21        & 3.184       \\\bottomrule       
\end{tabular}
}
\label{table1}
\end{table*}

To solve this non-linear optimization problem, as depicted in Equ. \ref{equ7}. We employ Gauss-Newton method \cite{bhardwaj2020differentiable} that iterative refine the planning variable $\tau$ based on $\hat{\mathbf{Y}}^\mathbf{T}_{1:T}$ as initial value. 

\section{Experiments}
\subsection{Experimental Setup}
We utilize the Waymo Open Motion Dataset (WOMD) \cite{ettinger2021large}, which contains over 500,000 samples that cover a wide range of real-world driving scenarios. It captures the complex interactions and dynamics among various types of traffic agents, including vehicles, cyclists, and pedestrians. For scene context $\mathcal{S}$ as inputs at the first stage, the historical states for each actor are sampled at 10Hz for the past one second ($T_h=10$), while the prediction and planning objectives are over the future 5 seconds at 1Hz for prediction and 10Hz for planning ($T=50$). The BEV image resolution for input and output is $H,W=128$, rendering $n=1.6$ pixels per meter in the real world. We set the hidden dimension as 96. For vector inputs, we maintain a maximum of $n_x=31$ surrounding actors sorted by their average distance to the AV. Following the standard measure, we set the anchors $k=6$ in multi-modal planning decoding. For the second stage of OPGP, the reference route $\mathcal{I}$ is smoothly intersected every 0.1m, and the size of transformed occupancy prediction $\mathbf{\hat{O}}^{\mathbf{T}}_{t}$ is $S=1000$, $D=20$. For planning purposes, we randomly split the recorded 20 seconds of WOMD, resulting in 297,669 samples for training and selected scenarios of interest for validation and testing (47,728 each).

\textbf{Metrics:} To ensure a comprehensive evaluation of OPGP performance, for occupancy predictions, we adopted the standard metrics proposed in challenges \cite{mahjourian2022occupancy}. The prediction metrics include \textbf{AUC} and \textbf{sIOU} for $\mathbf{\hat{O}}_{1:T}$. For planning metrics, we focus on the driving performance concerning human-likeness, motion capabilities, and safety during testing.

\textbf{Baselines:} To validate the prediction and planning performance of the OPGP framework, we compare it with several strong baselines. \textbf{1) For occupancy prediction}, we provide the vanilla STrajNet \cite{liu2022strajnet} without a planning head and VectorFlow-SwinT, the improved baseline predictor with simplified aggregation and encoding using the Swin Transformer bottleneck \cite{huang2022vectorflow}. \textbf{2) For open-loop planning}, we compare it with Vanilla IL, i.e., only the first stage of OPGP, and DIM \cite{rhinehart2018deep} devised with OPGP bottlenecks. These baselines also serve as ablative methods without some of the proposed modules in OPGP.

\subsection{Implementation Details}
We select the GELU activation function and apply a dropout rate of 0.1 after each layer to mitigate overfitting issues. Due to the large volume of data, we employ a distributed strategy on four NVIDIA A100 GPUs with a total batch size of 64. The AdamW optimizer is used with an initial learning rate of 1e-4, and a cosine annealing learning rate strategy is employed. The total number of training epochs is set to 30.

\subsection{Quantitative Results}
\textbf{1) Prediction Performance:} We report the testing performance of OPGP in comparison with other strong baselines. As shown in Table \ref{table2}, when compared with the Vanilla STrajNet, which is specifically designed for the occupancy prediction task, the prediction performance of OPGP exhibits a slight decrease. This is due to OPGP being a multi-task learning pipeline with both prediction and planning heads. More importantly, the current design only considers feature guidance from prediction towards planning. Consequently, the training objectives for multi-modal planning require further contributions to the decoding head for occupancy prediction.

\begin{table}[htp]
\centering
\vspace{-0.2cm}
\caption{Testing Results on Occupancy Predictions}
\begin{tabular}{l|llll}
\toprule
Models                               & Vec-AUC           & Vec-sIOU          & Occ-AUC           & Occ-sIOU \\\midrule
STrajNet\cite{liu2022strajnet}       & \textbf{0.856}     & \textbf{0.696}    & \textbf{0.146}   &  \textbf{0.023}    \\
VectorFlow\cite{huang2022vectorflow} & 0.813                & 0.656         & 0.112                & 0.017    \\\midrule
\textbf{OPGP}                        & 0.854                & 0.679         & 0.128             & 0.02    \\\bottomrule
\end{tabular}
\label{table2}
\end{table}

\textbf{2) Planning Performance:} We conduct open-loop planning testing. As shown in Table \ref{table1}, Vanilla IL, which employs the first stage of OPGP, achieves the lowest planning errors, demonstrating the performance gain of multi-modal configurations using $k$ anchors. In contrast, a uni-modal Gaussian for imitative planning in DIM cannot resolve the uncertainty in complex scenarios. The proposed OPGP yields low planning errors while enhancing safety compared to IL. The collision and off-route rates are significantly improved due to the occupancy prediction-guided optimization in the second stage. The well-designed cost function also promotes driving comfort by reducing jerks in motion metrics.

\begin{figure*}[ht]
    \centering
    \includegraphics[width=\linewidth]{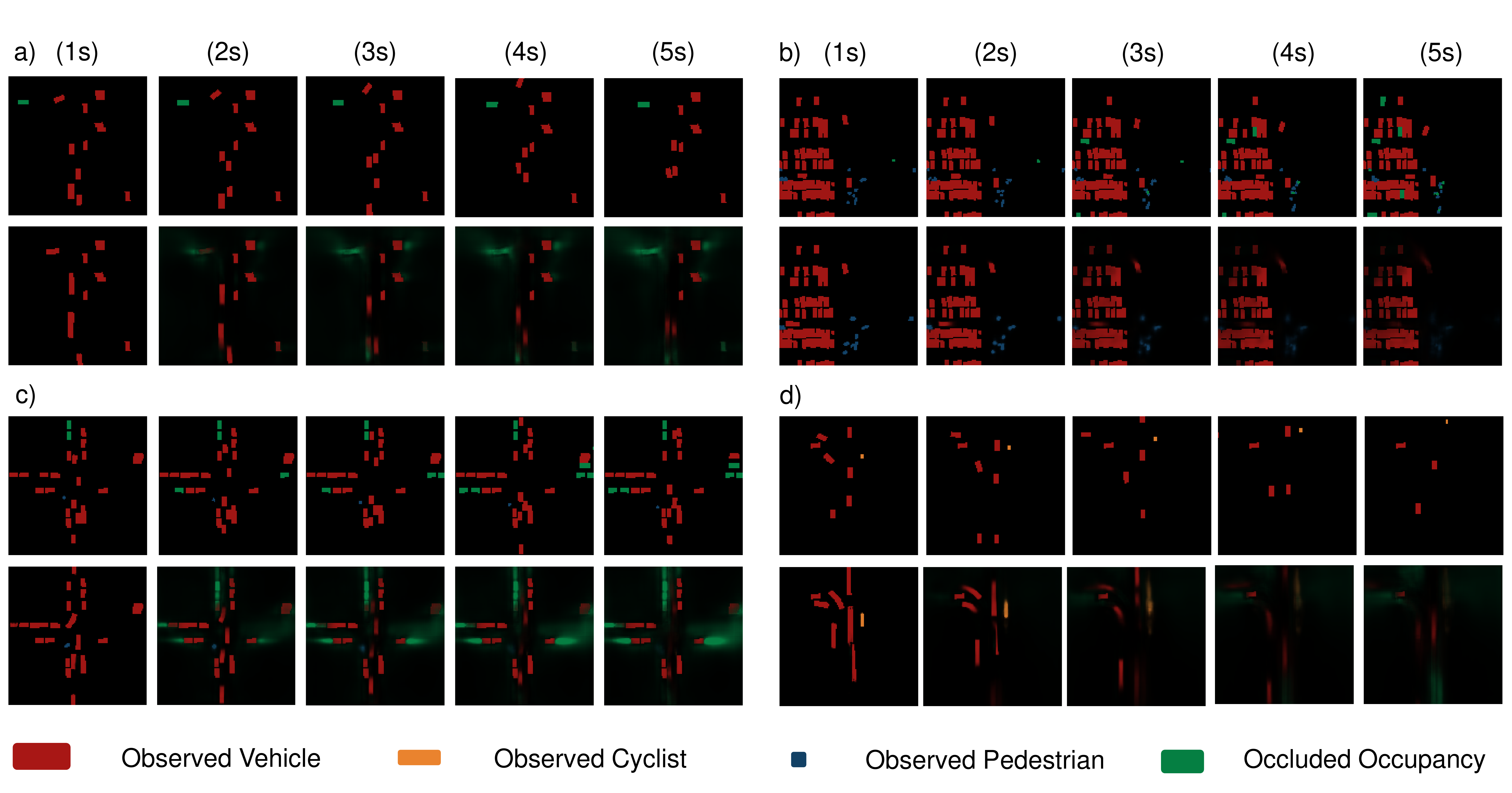}
    \caption{Qualitative results of occupancy predictions from the testing set. The sub-figures display several selected driving scenarios with specific features: a) T-intersections with merging; b) Parking lots with numerous pedestrians; c) Occluded intersections; d) Fork roads with adjacent cyclists.} 
    \label{fig3}
    \vspace{-0.2cm}
\end{figure*}

\subsection{Qualitative Results}
To gain a better understanding of the effectiveness of our OPGP, we first employ visualizations of occupancy prediction to assess its performance on multiple representative driving scenarios during testing. As shown in Fig. \ref{fig3}, the proposed method effectively handles both observed actors (a, c, d) and high-occlusion risk scenarios (a, c). Prediction accuracy is well-maintained for vehicles, but current methods can only ensure 3s prediction performance for smaller actors (b, d). To demonstrate the prediction-guided capability for planning, we display the planning result with corresponding transformed occupancy predictions.

\begin{figure}[htp]
    \centering
    \includegraphics[width=\linewidth]{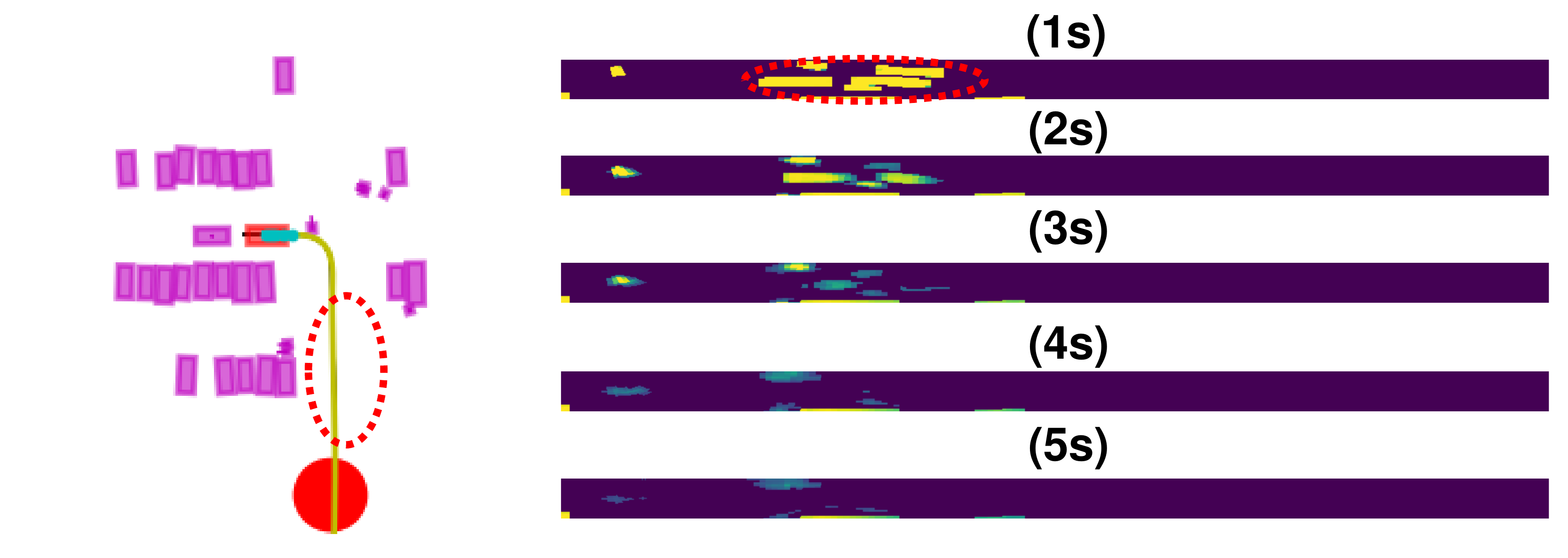}
    \caption{An example of refined planning results with corresponding transformed occupancy prediction for guidance. The AV (red) stops due to the approaching pedestrians. The prediction results circled in red dots refer to undetected actors in agent-centric methods.} 
    \label{fig4}
\end{figure}

As illustrated in Fig. \ref{fig4}, the planning rendered in cyan causes the AV (red) to stop due to the predicted approaching pedestrians. Moreover, as indicated by the red dotted circle, the occupancy prediction in OPGP can predict undetected actors that would be missed by agent-wise methods.

\subsection{Limitations and Future Work}
Although the current two-stage OPGP shows promising results, it also raises some topics that require further improvements. The first issue is the slight drop in occupancy prediction performance in multi-task learning. This requires joint interactions from network design to the learning scheme for prediction and planning. Another issue is the costly tuning for handcrafted weights and conservative prediction guidance, which forces us to manually define a safety distance. Occupancy prediction requires additional guidance in eliminating intractable pixels. Future work will aim to resolve these issues by improving network structures based on OPGP pipelines.

\section{Conclusions}
In this paper, we present a two-stage Occupancy Prediction-Guided Neural Planner (OPGP) that refines learning-based planning through prediction guidance in a joint manner. We develop an integrated learning-based framework with Transformer backbones, designed for comprehensive occupancy predictions and multi-modal planning objectives. Following the first stage outputs for prediction and planning, a transformed occupancy-guided optimization, built upon a curvilinear frame, achieves direct planning refinement through the use of handcrafted cost function designs. The prediction and planning performance are extensively validated using large-scale, real-world datasets (WOMD). Exhibiting robust performance in comparison to vanilla strong prediction baselines, the planning results demonstrate enhanced safety and driving smoothness. Furthermore, qualitative results substantiate the effectiveness of the transformed occupancy prediction guidance, revealing increased scalability in handling undetected and occluded actors when compared to agent-wise methods.

\bibliographystyle{IEEEtran}
\bibliography{b1}
\end{document}